# Approximate Piecewise Constant Curvature Equivalent Model and Their Application to Continuum Robot Configuration Estimation

Hao Cheng, Houde Liu*, Xueqian Wang* and Bin Liang

*Abstract*— The continuum robot has attracted more attention for its flexibility. Continuum robot kinematics models are the basis for further perception, planning, and control. The design and research of continuum robots are usually based on the assumption of piecewise constant curvature (PCC). However, due to the influence of friction, etc., the actual motion of the continuum robot is approximate piecewise constant curvature (APCC). To address this, we present a kinematic equivalent model for continuum robots, i.e. APCC 2L-5R. Using classical rigid linkages to replace the original model in kinematic, the APCC 2L-5R model effectively reduces complexity and improves numerical stability. Furthermore, based on the model, the configuration self-estimation of the continuum robot is realized by monocular cameras installed at the end of each approximate constant curvature segment. The potential of APCC 2L-5R in perception, planning, and control of continuum robots remains to be explored.

## I. INTRODUCTION

In recent years, the research on the planning and control of continuum robots has been gradually focused. In terms of disaster relief, nuclear and radiation equipment maintenance, toxic waste sampling, pipeline monitoring, etc., it is not suitable for people or large equipment to enter due to the small space and great danger. Therefore, the continuum robot has become an essential choice for its slim body and flexible movement [1]. Continuum robots have an excellent ability to bend and avoid obstacles, and can change their shapes to adapt to the environment and overcome the limitations of various obstacles. It is widely used in the special occasions of autonomous operation under the unstructured environment, such as medical treatment, military, disaster relief, ocean exploration and other fields. However, due to its redundancy of degrees of freedom, the perception, planning, and control of continuum robots are still in the research and exploration stage. Among them, continuum robot environment perception (i.e. configuration estimation and environmental map construction) is the basis to realize its effective planning and control in a complex and unknown environment, which could avoid obstacles and complete tasks. In other words, it is necessary to solve the problem that it can obtain the state of robots in various environments in real-time. To realize the continuum robot environment perception, the kinematic relationship of the continuum robot is urgent to be modeled and described.

\* This work was supported by National Natural Science Foundation of China (61803221 & U1813216), Guangdong Young Talent with Scientific and Technological Innovation (2019TQ05Z111), and the Interdisciplinary Research Project of Graduate School of Shenzhen of Tsinghua University (JC2017005).

Hao Cheng, Houde Liu (corresponding author), and Xueqian Wang (corresponding author) are with the Shenzhen International Graduate School, Tsinghua University, 518055 Shenzhen, China (E-mails: Liu.hd@sz.tsinghua.edu.cn and wang.xq@sz.tsinghua.edu.cn).
Bin Liang is with the Department of Automation, Tsinghua University, 100084 Beijing, China.

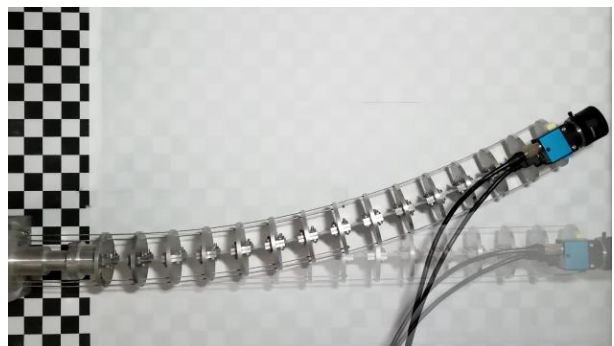

Figure 1. The prototype of the continuum robot.

In favor of engineering implementation, structure design and application of continuum robot are often based on Piecewise Constant Curvature (PCC) assumption. Based on the PCC assumption, the continuum robot is composed of constant curvatures (that can change with time), to reduce the application difficulty of the continuum robot effectively. However, due to the influence of friction and other factors in engineering implementation, the ideal PCC assumption is usually challenging to be satisfied. Instead, it is approximate to the configuration of the PCC assumption within a certain deviation range. Besides, the existing kinematic model of continuum robot based on PCC assumption is prone to numerical instability at the singular points and has strong nonlinear, which is not conducive to the realization of continuum robot environment perception. Therefore, it is necessary to improve the kinematics model of the continuum robot under the Approximate Piecewise Constant Curvature (APCC) assumption, to effectively avoid the influence and limitation of the PCC assumption deviation, numerical instability and nonlinearity.

On the other hand, with the development of simultaneous localization and mapping (SLAM) technology, which has been widely used in mobile robots, e.g. unmanned aerial vehicles (UAV) and mobile cars. Meanwhile, with the development of visual SLAM technology, the advantages of low cost and abundant information of cameras are brought into localization and mapping, such as [2, 3]. At present, the localization of visual SLAM technology is mainly based on the use of the visual odometer of mobile robots. To the best of our knowledge, this is the first time that SLAM is introduced into such a robotic manipulator as a continuum robot to realize real-time configuration estimation. It is mainly because the classical rigid robot manipulator can be accurately modeled by D-H parameters and other methods, while for non-rigid robots such as continuum robots and soft robots, it is challenging to estimate its configuration in real-time. Visual SLAM technology provides a new way for real-time configuration estimation of continuum robots.

In this work, we propose an approximate piecewise constant curvature kinematics equivalent model, i.e. APCC 2L-5R model. More specifically, two classical rigid linkages are connected by five revolute joints, and their kinematic relations are constrained by specific parameters (some of which are determined by calibration), which are equivalent to the kinematic relations of approximate piecewise constant curvature segment (closer to the actual motion) within an acceptable error range.

This work contributes:

- Propose an approximate piecewise constant curvature kinematics equivalent model, i.e. APCC 2L-5R (2 linkages & 5 revolute joints), which can model the motion of actual continuum robot more accurately and reduce the complexity of the model.
- Provide a calibration method for the deviation between the approximate piecewise constant curvature and the ideal piecewise constant curvature of a real continuum robot.
- Based on the APCC 2L-5R model, a generalized epi-polar constraint relation was established for the planar continuum robot, and the algorithm was validated by simulations and experiments.

The outline of this paper is shown as follows: In section II, we review further related work. In section III, we introduce the APCC 2L-5R model and deviation calibration method, so that the approximate constant curvature segment (ACC-segment) of APCC continuum robot can be equivalent through the classical rigid body. In section IV, based on the APCC 2L-5R model, the configuration of continuum robots can be estimated by using generalized epi-polar constraint. Section V presents results on simulated and real data. Section VI summarizes the results and describes future work.

## II. RELATED WORK

The continuum robot was proposed in 1999; Robinson et al. distinguished it as a discrete robot, serpentine robot, and continuum robot according to its structural characteristics [1]. The research of continuum robots mainly comes from bionic design. Walker et al. designed an elephant-nosed robot that can bend and grab objects around [4]. The continuum robot has promising applications, e.g. minimally invasive surgery [5], nuclear power plant exploration [6]. Continuum robot has several types, such as soft robot [10], cable-driven continuum robot [7], etc.

Since continuum robots are deformable, they take on shapes that are general curves in space, and accurate control is a hard task. Due to the actual needs of engineering implementation, the structural design, perception, planning, and control of continuum robots need to follow some assumptions, among which the piecewise constant curvature assumption is widely used [8,9]. However, as the influence of friction, etc., the motion of the actual continuum robot is approximately piecewise constant curvature [11]. R. k. Katzschmann et al. established a dynamic equivalent model for the soft robot based on the assumption of PCC, and verified that the soft robot could be controlled based on the assumption of PCC in the presence of deviations through experiments, but the accuracy has an impact [10].

Perception of continuum robots in real-time is a necessary condition for accurate and robust control. For the real-time perception of continuum robots, there are two main approaches: (1) force perception method based on the mechanical model, which depends on a specific continuum robot model, and there is no universal solution. S. Huang et al. made a mechanical analysis of the continuum robot by establishing a pseudo-rigid body model, and completed the estimation of the end position of the continuum manipulator [11, 12]. (2) based on sensors, real-time perception can be achieved mainly through the installation of sensors such as camera, IMU or optical fiber, and its perception method can be universal. For the vision sensor, M. W. Hannan et al. estimated the curvature of each segment from the external global image of the manipulator based on computer vision technology to obtain the overall shape [13]; A. Reiter et al. proposed a learn-based multi-segment continuum robot configuration estimation method [14]; For other sensors, S. Sareh et al. completed the configuration measurement through the optical fiber installed on the continuum robot [15]. For the advantages of low cost and abundant information, the vision sensor has great potential. However, at present, external camera information is mainly used based on visual methods, which limits its application in unknown unstructured scenes. Therefore, visual SLAM technology can be introduced to realize real-time configuration self-estimation of continuum robots.

R. Pless first derives the generalized epi-polar constraint through Plücker vectors and applies it to multi-camera systems [16]. L. Kneip et al. studied the problem of multiple camera systems [17]. X. Peng et al. proposed articulated multi-perspective cameras model extends the case of multiple camera systems from relative immobilization between cameras to articulated confinement, and applies it to truck motion estimation [18].

In this work, we propose a continuum robot kinematics model based on approximate piecewise constant curvature equivalent to the classical rigid robot kinematics. On this basis, APCC 2L-5R is applied to the configuration self-estimation of planar continuum robots by the generalized epi-polar constraint theory.

## III. APCC 2L-5R KINEMATIC EQUIVALENCE MODEL

Before we introduce the APCC 2L-5R kinematic equivalence model, it is worth reviewing the piecewise constant curvature (PCC) assumption of continuum robots. Continuum robot under ideal PCC is composed of a series of arcs which curvature is a certain value. However, there is a deviation between the actual continuum robot and the ideal PCC due to the influence of friction in engineering implementation. Our approach is improved from PCC model. The actual motion of the continuum robot can be described more accurately, and the kinematics is equivalent to the classical rigid manipulator, which can effectively reduce the complexity of the model. Next, we will review the PCC and introduce the APCC 2L-5R model. Finally, a deviation calibration method is given.

### A. Review Piece Constant Curvature Kinematics

In the design and research of the continuum robot, a finite degree of freedom model is established to describe its motion based on the assumption of PCC. Based on PCC and fixed length, the continuum robot can be divided into multiple constant curvature segments (CC-segment), as shown in Fig.2 (a). The geometric relationship of its kinematic is shown in Fig. 2 (b).

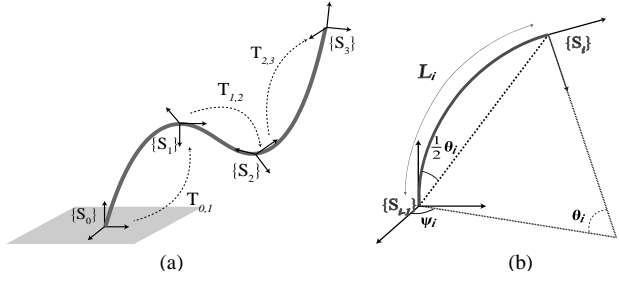

(a)   (b)

Figure 2.   The geometric relationship of PCC continuum robot. (a) An example of an overall PCC continuum robot. (b) shows the geometric relationship of a CC-segment.

The kinematic homogeneous transformation relation as follows:

$$\mathbf{T}_{i-1,i}(\theta_i) = \mathbf{R}_z(\psi_i) \begin{bmatrix} \mathbf{R}_x(\theta_i) & \frac{L_i}{\theta_i}\begin{bmatrix} \cos(\theta_i) - 1 \\ \sin(\theta_i) \\ 0 \end{bmatrix} \\ 0 & 1 \end{bmatrix} \mathbf{R}_z(-\psi_i). \quad (1)$$

Where, $L_i$, a certain value, is the length of the $i$-th CC-segment; $\psi_i$ as the axial rotation angle, $\theta_i$ for axial deflection angle.

$\psi_i$ and $\theta_i$ determine continuum robot pose transformation relationship between the CC-segments. However, since the above formula contains a term with the state variable as the denominator, i.e. singular point ($\theta_i = 0$), there is numerical instability. In addition, although the continuum robot satisfies the assumption of PCC in the ideal state through structural design, the actual continuum robot just approximately satisfies PCC.

*B. The APCC 2L-5R*

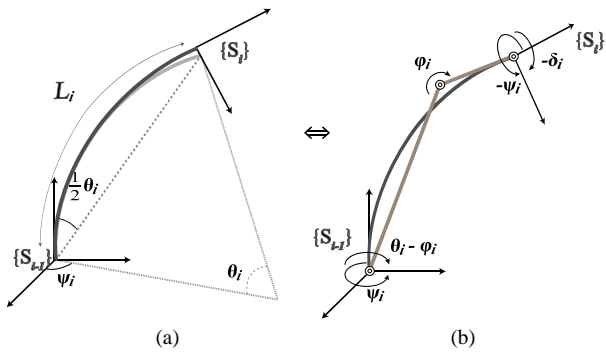

(a)   (b)

Figure 3.   APCC 2L-5R Equivalent model diagram. (a) An ACC-segment of overall APCC continuum robot, the grey arc is the shape of a segment of ideal PCC continuum robot. The black arc is the shape of segments of the actual continuum robot. (b) shows the connection of the equivalent model.

In this section, we propose a kinematic equivalent model of continuum robot under APCC, i.e. APCC 2L-5R model. The idea of this method is to use five revolute joints to connect two classical rigid linkages, and to constrain the kinematic relationship of the two linkages equivalent to the ACC-segment in the acceptable error range according to the specific parameters (some of which are determined by calibration). The whole model is composed of a series of multiple ACC-segments.

Fig.3 illustrates the principle that APCC 2L-5R model is equivalent to actual ACC-segment. The emphasis of this method lies in the combination structure of 2L-5R, i.e. the two linkages - the five revolute joints, and its parameter relationship, especially the relationship between axial deflection angle $\theta$, $\psi$.

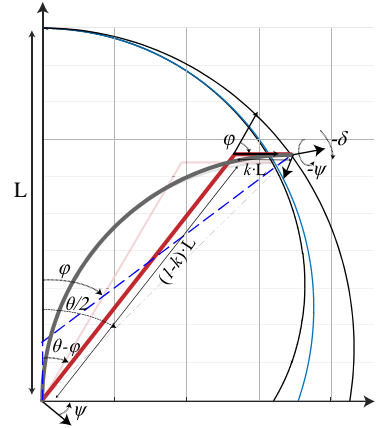

Figure 4.   Diagram of the deflection plane of APCC 2L-5R. The red line is a linkage of APCC 2L-5R model. The inside grey arc is one ACC-segment. The outside arcs are the end trajectories of APCC, among which the right arc is the trajectories of ideal PCC.

Fig. 4 shows the equivalent principle on the axial deflection plane. The equivalent model is composed of two linkages and five revolute joints. The length of the original ACC-segment determines the length of rigid linkages. There are constraint relationships in the five revolute joints. According to the geometric relationship in Fig. 4, there are two symmetric equivalent methods (blue dotted line and solid red line). In practice, the red one is often used to realize the equivalent method, while the blue one is only used to derive the geometric constraint relationship between equivalent deflection angles, i.e. $\theta$ and $\varphi$. The following constraints are obtained according to the geometric relationship of the blue one.

$$(1-k)L\sin(\varphi) = \big((1-k)L\cos(\varphi) + kL\big)\tan\big(\tfrac{\theta}{2}\big). \quad (2)$$

Further simplification results in:

$$\tan\left(\frac{\theta}{2}\right) = \frac{\sin(\varphi)}{\cos(\varphi) + \frac{k}{1-k}} = \frac{\sin(\theta)}{\cos(\theta) + 1} = \frac{1-\cos(\theta)}{\sin(\theta)}. \quad (3)$$

The specific parameters in the equivalent method are shown in table 1.

The deflection angle at the end of the ACC-segment can be modified as follows:

$$\hat{\theta} = \theta - \delta. \quad (4)$$

Among them,

$$\delta = \delta_{max} \cdot \frac{\theta}{\theta_{max}} \quad (5)$$

Among the above parameters, $k$ and $\delta_{max}$ shall be determined by calibration.

TABLE I. THE PARAMETERS OF APCC 2L-5R MODEL

| Symbol | Definition | Property |
|---|---|---|
| $\theta$ | The ideal axial deflection angle of one ACC-segment | if no deviation, the tip deflection angle |
| $\psi$ | Axial rotation angle | Configuration variable |
| $\hat{\theta}$ | The modified estimator of axial deflection angle | Equivalent model state variables |
| $\delta$ | Correction angle of deflection | See Eq. (5) |
| $\delta_{max}$ | The maximum deviation of the deflection angle | Need to calibrate |
| $k$ | Deviation coefficient | |
| $L$ | The length of one ACC-segment | Constant |
| $l_1$ | The length of the first rigid linkage | $l_1 = (1-k) \cdot L$ |
| $l_2$ | The length of the second rigid linkage | $l_2 = k \cdot L$ |
| $\theta - \varphi$ | The first rotation angle in APCC 2L-5R model | See Eq. (3) |
| $\varphi$ | The second rotation angle in APCC 2L-5R model | |

In this method, the deviation coefficient $k$ is determined to be a constant through calibration, i.e. the only approximate end trajectory is determined after calibration. It can be seen in Fig. 4 that when the axial deflection angle is too large, the deviation increases too much, so that the equivalent model is difficult to describe the actual situation. However, in the engineering implementation, for material restrictions, the axial deflection angle of each ACC-segment is limited, generally less than 60°, hence the equivalent model can work effectively. In practical engineering applications, the kinematic equivalent approach can effectively reduce the complexity, and the approximate deviation correction can be considered to ensure the accuracy of the model.

*C. Calibration Method for APCC*

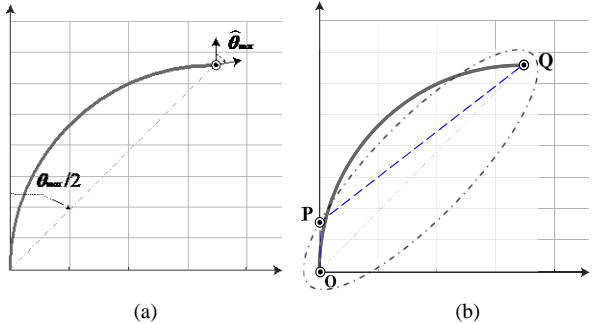

Figure 5. Deviation calibration diagram. (a) Calibrate the maximum deviation angle. (b) Calibrate deviation coefficient.

In engineering implementation of continuum robot based on PCC, there is a certain deviation between the actual motion and the ideal PCC motion, which depends on the mechanical structure, material properties and friction coefficient, etc.; hence it is necessary to calibrate the deviation at first. We propose a calibration method for the deviation, which can be used to calibrate the deviation of a continuum robot in advance.

For a continuum robot based on PCC design, it is composed of several CC-segments, and each segment needs to be calibrated separately. The purpose of calibration is to calibrate the maximum deviation angle $\delta_{max}$ and deviation coefficient $k$.

The calibration method steps are as follows:

**Step 1:** Choose an uncalibrated ACC-segment of the continuum robot, then axial rotation angle $\psi$ adjusts to 0°, axial deflection angle $\theta$ to $\theta_{max}$;

**Step 2:** Make the camera parallel to the plane formed by the segment, and carry out visual sampling on the segment to obtain the geometric relation sampling figure, as shown in Fig. 5 (a). The coordinate of the front (near the root) of the segment is set as the coordinate origin;

**Step 3:** Calibrate the maximum deviation angle. Measure the angle $\theta_{max}/2$ and $\hat{\theta}_{max}$ in the sampling figure and get the maximum deviation angle $\delta_{max} = \theta_{max} - \hat{\theta}_{max}$;

**Step 4:** Calibrate deviation coefficient. As shown in Fig. 5(b), with the origin and the endpoint as the focus, the length of the long axis is the length of the constant curvature segment $L$, make an ellipse. The ellipse intersects the positive half axis of the $y$-axis at point **P**, then the deviation coefficient is $k = \overline{OP}/L$.

**Step 5:** Repeat the above steps until the calibration of each ACC-segment of the whole continuum robot is completed.

Next, we present an application of the APCC 2L-5R model: configuration estimation for a planar continuum robot.

## IV. APCC 2L-5R APPLIED TO PLANAR CONTINUUM ROBOT CONFIGURATION ESTIMATION

Based on APCC 2L-5R, we derive APCC Generalized Epi-polar Constraint. By placing monocular cameras at the end of each ACC-segment and solving the APCC Generalized Epi-polar Constraint, the configuration estimation was performed.

We first review the concept of generalized polar constraint, then derive APCC generalized epi-polar constraint, and finally introduce the 1-point solver.

*A. Review of Generalized Epi-polar Constraint*

The generalized epi-polar constraint is an extension of the epi-polar constraint, which from the research of generalized camera model (multi-camera system) [20]. Its characteristic is measurement rays do no longer intersect in a common point, while regular camera intersects in a common point, i.e. focal point.

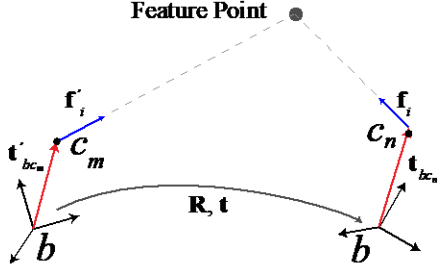

Figure 6. A case of generalized epi-polar constraint.

Fig. 6 shows a typical generalized epi-polar constraint case. The generalized epi-polar constraint relation as follows:

$$\begin{bmatrix} \mathbf{f}_i \\ \mathbf{t}_{bc_n} \times \mathbf{f}_i \end{bmatrix}^T \begin{bmatrix} \mathbf{E} & \mathbf{R} \\ \mathbf{R} & 0 \end{bmatrix} \begin{bmatrix} \mathbf{f}_i' \\ \mathbf{t}_{bc_m}' \times \mathbf{f}_i' \end{bmatrix} = 0. \quad (6)$$

Where $\{\mathbf{f}_i, \mathbf{f}_i'\}$ are corresponding direction vectors based on the multi-camera system center frame. $\mathbf{t}_{bc,i}$ and $\mathbf{t}_{bc,i}'$ as the position of the camera relative to the center in two views. $\mathbf{E} = [\mathbf{t}]_\times \mathbf{R}$ is essential matrix. $\mathbf{R}$ and $\mathbf{t}$ are transformation parameters between the two viewpoints.

When $\mathbf{t}_{bc,i}$ and $\mathbf{t}_{bc,i}'$ are zero, (6) is reduced to the common epi-polar constraint.

### B. Generalized Epi-polar Constraint on APCC 2L-5R

In recent studies, the multi-camera system was extended from the relatively fixed to the relatively variable situation between cameras [18, 19]. In this work further introduces it into the variable APCC 2L-5R, and obtains APCC generalized epi-polar constraint. Through this constraint relationship, configuration estimation of planar continuum robot can be performed.

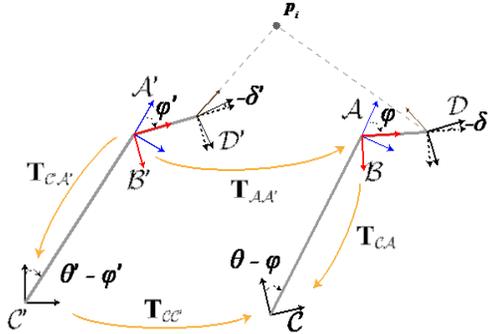

Figure 7. The illustration of generalized epi-polar constraint on APCC 2L-5R model. There are four degrees of freedom, i.e. root (frame C) movement and the change of ACC-segment configuration. External feature points correspondences of the two views can be observed by the camera configured at the end (frame D).

For any CC - segment of planar cases APCC 2L- 5R model, the number of configurations is 1, i.e. axial deflection angle $\psi = 0$). Fig. 7 is an example of APCC 2L-5R model in the two views before and after the movement. Let $\mathbf{t}_{BD}$ be the position of camera inside the reference frame B. Let $\{\mathbf{f}_i, \mathbf{f}_i'\}$ be feature corresponding direction vectors. Let $\mathbf{R}_{BB'}$ and $\mathbf{t}_{BB'}$ be the transformation variables from B' to B. According to equation (4), generalized epi-polar constraint in B as

$$\begin{bmatrix} \mathbf{R}_Z(-\delta)\mathbf{f}_i \\ \mathbf{t}_{BD} \times \mathbf{R}_Z(-\delta)\mathbf{f}_i \end{bmatrix}^T \begin{bmatrix} [\mathbf{t}_{BB'}]_\times \mathbf{R}_{BB'} & \mathbf{R}_{BB'} \\ \mathbf{R}_{BB'} & 0 \end{bmatrix} \begin{bmatrix} \mathbf{R}_Z(-\delta')\mathbf{f}_i' \\ \mathbf{t}_{BD}' \times \mathbf{R}_Z(-\delta')\mathbf{f}_i' \end{bmatrix} = 0. \quad (7)$$

In order to estimate the configuration of the continuum robot, we need to consider the position relationship between two rigid linkages. From the angle relation in APCC 2L-5R model, the following constraint relations can be deduced:

$$\mathbf{T}_{AA'}\mathbf{T}_{A'B'} = \mathbf{T}_{AB}\mathbf{T}_{BB'}$$
$$\mathbf{T}_{CA}\mathbf{T}_{AA'} = \mathbf{T}_{CC'}\mathbf{T}_{C'A'} \quad (8)$$

Then the transformation from A' to A, $\mathbf{R}_{AA'}$ and $\mathbf{t}_{AA'}$ as

$$\begin{cases} \mathbf{R}_{AA'} = \mathbf{R}_{CA}^T \mathbf{R}_{CC'} \mathbf{R}_{C'A'} \\ \mathbf{t}_{AA'} = \mathbf{R}_{CA}^T \mathbf{t}_{CC'} + \mathbf{R}_{AA'} \mathbf{t}_{C'A'} - \mathbf{t}_{CA} \end{cases} \quad (9)$$

And

$$\mathbf{T}_{AB} = \begin{bmatrix} \mathbf{R}_Z(\varphi) & 0 \\ 0 & 1 \end{bmatrix}$$

Meanwhile, the transformation from B' to B, $\mathbf{R}_{BB'}$ and $\mathbf{t}_{BB'}$ as

$$\begin{cases} \mathbf{R}_{BB'} = \mathbf{R}_z^T(\varphi) \mathbf{R}_{AA'} \mathbf{R}_z(\varphi') \\ \mathbf{t}_{BB'} = \mathbf{R}_z^T(\varphi) \mathbf{t}_{AA'} \end{cases} \quad (10)$$

By substituting (10) into (7), then

$$\begin{bmatrix} \mathbf{R}_Z(\varphi)\mathbf{R}_Z(-\delta)\mathbf{f}_i \\ \mathbf{R}_Z(\varphi)\mathbf{t}_{BD} \times \mathbf{R}_Z(-\delta)\mathbf{f}_i \end{bmatrix}^T \begin{bmatrix} [\mathbf{t}_{AA'}]_\times \mathbf{R}_{AA'} & \mathbf{R}_{AA'} \\ \mathbf{R}_{AA'} & 0 \end{bmatrix}$$
$$\begin{bmatrix} \mathbf{R}_Z(\varphi')\mathbf{R}_Z(-\delta')\mathbf{f}_i' \\ \mathbf{R}_Z(\varphi')\mathbf{t}_{BD}' \times \mathbf{R}_Z(-\delta')\mathbf{f}_i' \end{bmatrix} = 0. \quad (11)$$

The above equation establishes the relationship between the intermediate joints. It is similar to articulated multi-perspective cameras (AMPC) in [18].

Next, consider the transformation between C and A. Through geometric relationship, $\mathbf{R}_{CA}$ and $\mathbf{t}_{CA}$ as follows:

$$\begin{cases} \mathbf{R}_{CA} = \mathbf{R}_z(\theta - \varphi) \\ \mathbf{t}_{CA}^T = \frac{k}{1-k} L [\sin(\theta - \varphi) \quad \cos(\theta - \varphi) \quad 0] \end{cases} \quad (12)$$

In addition, from (3), we obtain

$$\begin{cases} \sin(\theta - \varphi) = \sin(\varphi) - \frac{k}{1-k}\sin(\theta) \\ \cos(\theta - \varphi) = \cos(\varphi) - \frac{k}{1-k}\cos(\theta) + \frac{k}{1-k} \end{cases} \quad (13)$$

Substituting (9) and (12) into (11), the complete generalized epi-polar constraint on APCC 2L-5R is obtained

$$\begin{bmatrix} \mathbf{R}_Z(\theta)\mathbf{R}_Z(-\delta)\mathbf{f}_i \\ \mathbf{R}_Z(\theta)\mathbf{t}_{BD} \times \mathbf{R}_Z(-\delta)\mathbf{f}_i \end{bmatrix}^T \begin{bmatrix} \mathbf{E}_{APCC} & \mathbf{R}_{CC'} \\ \mathbf{R}_{CC'} & 0 \end{bmatrix}$$
$$\begin{bmatrix} \mathbf{R}_Z(\theta')\mathbf{R}_Z(-\delta')\mathbf{f}_i' \\ \mathbf{R}_Z(\theta')\mathbf{t}_{BD}' \times \mathbf{R}_Z(-\delta')\mathbf{f}_i' \end{bmatrix} = 0. \quad (14)$$

$$\mathbf{E}_{APCC} = [\mathbf{t}_{CC'}]_\times \mathbf{R}_{CC'} + \mathbf{R}_{CC'}[\mathbf{t}_{C'A'}]_\times - [\mathbf{t}_{CA}]_\times \mathbf{R}_{CC'}$$

Based on the above formula, each ACC-segment is solved one by one from the root of the continuum robot, to estimate the configuration of the whole continuum robot. There is only one unknown variable $\theta$ ($\delta$ is function of $\theta$), hence it can be solved by a single correspondence.

### C. 1-Point Solver

In the above constraint equation (13), there is a nonlinear trigonometric function relationship, hence there is no analytic solution result; and the iterative solution is slow and greatly influenced by the initial value. By using the trigonometric identity, we can linearize it into higher order algebraic equations for solving, which can improve the solving speed and accuracy.

Introducing auxiliary variables:

$$\begin{cases} s_1 = \sin(\theta) \\ c_1 = \cos(\theta) \end{cases} \quad \begin{cases} s_2 = \sin(\varphi) \\ c_2 = \cos(\varphi) \end{cases} \quad (15)$$

Substituting (15) into (3), then

$$\begin{cases} s_1^2 + c_1^2 - 1 = 0 \\ s_2^2 + c_2^2 - 1 = 0 \\ s_2 c_1 - c_2 s_1 + s_2 - \dfrac{k}{1-k} s_1 = 0 \end{cases} \quad (16)$$

By adding auxiliary variables, the unknowns become four. One correspondence is substituted into (14) to obtain one algebraic equality relation; then together with the three algebraic equality relations of (16), constitutes a quaternion higher order algebraic equation system. By adding auxiliary variables, the unknowns become 4. One correspondence is substituted into (12) to obtain one algebraic equality relation; then together with the three algebraic equality relations of (14), constitutes a higher order algebraic equation system. K. Zuzana et al. provided an automatic generator of minimal problem solvers through algebraic geometry [21]. For this problem solver, the real-time solution can be realized within the range of milliseconds, so that realize the real-time configuration self-estimation of continuum robots based on vision. To be honest, algebraic geometry is a complex math problem, which is used to generate above solver. The solver can work effectively when the deviation (deviation coefficient $k$ and the maximum deviation of the deflection angle $\delta_{max}$) is small, i.e. $\delta_{max}$ is below 5°, which can be ignored. In addition, only if $k=0.25$, the solver gives a solution, which is a problem that needs further analysis. Fortunately, in practical application, if the deviation coefficient $k$ is approximately 0.25, estimation is effective using $k=0.25$.

## V. EXPERIMENTS

Now we verify our method by simulation and experiment. First, the calibration process is demonstrated through the actual continuum robot prototype. Next, the 1-Point solver is evaluated by simulation. Finally, the estimating algorithm validates by real data.

### A. Deviation calibration experiment

In this section, we present a real calibration process of one ACC-segment.

According to the calibration steps in III.C, we move the continuum robot and take the visual information when $\theta$ is 0 and $\theta_{max}$ as shown in Fig. 8. We compare the results of deviation calibration when the maximum deflection angle is different ($\theta_{max,1}$, $\theta_{max,2}$, $\theta_{max,3}$, $\theta_{max,4}$), and illustrate the effectiveness of APCC 2L-5R model to describe the actual continuum robot motion.

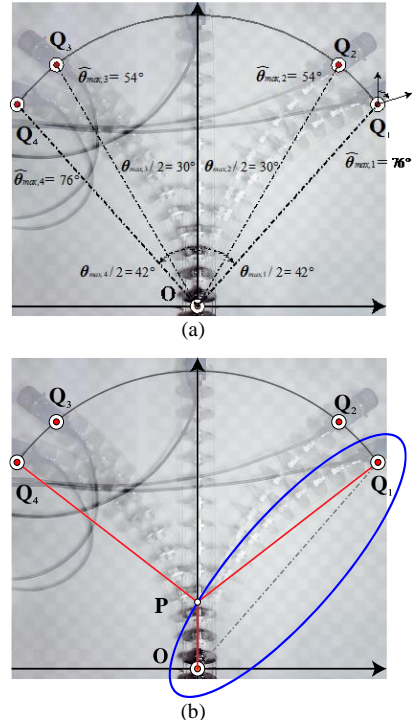

Figure 8. Deviation calibration experiment visual sampling and geometric analysis. (a) Calibrate the maximum deviation angle. (b) Calibrate deviation coefficient.

In above experiment, $\theta_{max,1}=\theta_{max,4}=84°$, $\theta_{max,2}=\theta_{max,4}=60°$. Therefore, we obtain that the maximum deviation angle $\delta_{max,1} = \theta_{max,1} - \hat{\theta}_{max,1} = 42 \times 2 - 76 = 8°$. In the same way, $\delta_{max,2} = \delta_{max,3} = 6°$, $\delta_{max,4} = \delta_{max,1} = 8°$, which accords with the structural symmetry of the planar continuum robot. The deviation coefficient of four positions ($\mathbf{Q}_1 \sim \mathbf{Q}_4$) are close: $k_1 = \overline{OP}_1 / L = 11.1/49.6 = 0.224$, $k_2 \approx k_3 \approx k_4 \approx 0.224$. $k$ is close to 0.25, so the motion is close to the ideal PCC.

The following conclusions can be drawn from the above calibration experiments.

1) Considering that the maximum deflection angle is $\theta_{max,1}$, the actual calibration deviation of the maximum deflection angle $\delta_{max,1} = 8°$. From (5), when $\theta = 60°$, $\delta=8\times60/84=5.7°$, which is close to the actual situation $\mathbf{Q}_2$. Therefore, the assumption of linear transformation of deflection angle deviation is applicable to the actual situation.

2) In this case, the planar continuum robot accords with the structural symmetry. Hence, the calibration results of $\mathbf{Q}_1$ & $\mathbf{Q}_4$, $\mathbf{Q}_2$ & $\mathbf{Q}_3$ are close, respectively. In some special case, the symmetry is not satisfied. We could calibrate in different deflection planes.

3) The deviation coefficient $k$ determines the end trajectory. In this case, the end trajectory corresponding to $k$ well depicts the actual end trajectory. The comparation of end trajectories shown in Fig. 9.

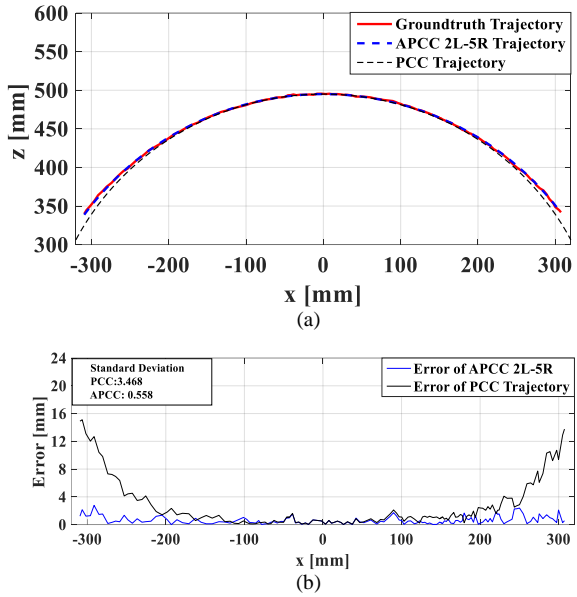

Figure 9. The comparison of different end trajectories. Through deviation calibration of the actual continuum robot, the end trajectory of APCC 2L-5R well describes the actual motion. With the increase of the deflection angle, the actual end trajectory deviated greatly from the ideal PCC trajectory. The bottom figure shows the error between APCC 2L-5R or PCC to the actual trajectory, the standard deviation of PCC error:3.468, APCC (ours): 0.558.

## B. Evaluation of the 1-Point Solver

In this section, we evaluate the 1-point solver in simulation. More specifically, the planar continuum robot moves in the ***x-o-y*** plane, which satisfies the PCC assumption. Let the length of each segment $L$=0.5m, and assume the focal length is 500 pixels. The axial deflection angle $\theta \in [-50°, 50°]$. The step of $\theta$ are generated randomly, and $|\theta - \theta'| < 5°$. The 1-Point solver is evaluated by two cases, fixed root and moved root. In the moved root case, we consider the rotation angle of root $\alpha \in [-5°, 5°]$, so the translation distance is under 0.1m. Trough adding Gaussian noise in observation, the error distribution was obtained by taking the average of 5000 measurements.

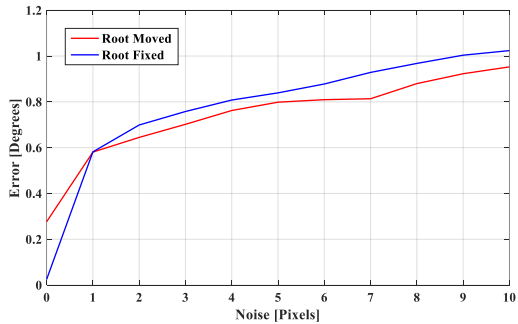

Figure 10. Noise resilience of the 1-point solver for the configuration angle $\theta$ in root moved and root fixed case, which caused by noise on the image points. Results are averaged over 5000 random experiments.

The error increases with the pixel noise but remains within a small range. From the evaluation, we know that the 1-Point solver is robust with the noise in observation.

## C. Experimental Validation

Next, we use the single ACC-segment planar continuum robot in the above deviation calibration experiment to verify the generalized epi-polar constraint. We constructed this planar continuum robot by elasticity steels skeleton, which is illustrated in Fig. 1. We installed a Daheng Imaging camera at the end of ACC-segment. The cameras have a resolution of 1202×964 and are equipped with $f$=16mm lenses. In order to test the validity of the generalized epi-polar constraint of APCC 2L-5R, we used the offline method to collect images, i.e. record images at some deflection angles. We obtain inner point feature correspondences using the RANSAC algorithm. For the limitation of solver, we consider the motion of this prototype as the ideal PCC. We recorded several images, and matched point features by extracting SURF features. In addition, we collected the global image through the industrial camera installed on the top of the planar continuum robot, thus enabling the collection of ground truth. Note that the purpose of the experiment is to verify the effectiveness of configuration estimation through solving the generalized epi-polar constraint (14), i.e. the front end of perception system; Without filtering or nonlinear optimization, i.e. the back end of perception system. The recorded images sequence consists of 21 images, with $\theta$ increasing linearly from 0 to 55 °.

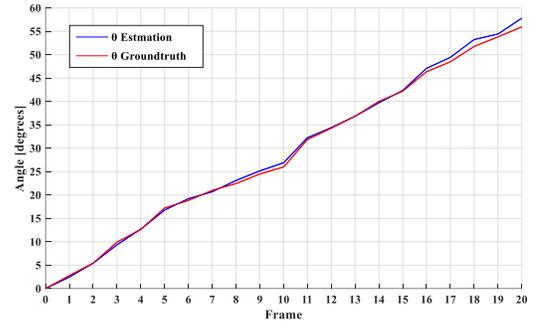

Figure 11. Configuration estmation results obtained by real data. The red one is ground truth $\theta$, and the blue one is the estimation result using our approach.

Fig. 11 illustrates the performance of our approach and compares it against ground truth. As can be observed, the configuration of planar continuum robots can be recovered through solving generalized epi-polar constraints based on APCC 2L-5R kinematics equivalent model. Though the single ACC-segment experiment, the result generally shows that our method can be applied to the real case. Note that the results on real data show that the errors accumulate with the estimation process. In order to realize the configuration estimation of the multi-segment planar continuum robot, filtering or nonlinear optimization should be used to mitigate the influence of errors on the configuration estimation of the subsequent ACC-segment. On the other hand, due to the theoretical limitation that the polar constraint cannot solve the pure rotation problem, the axial rotation angle $\psi$ cannot be estimated for the spatial continuum robot simply by solving the generalized epi-polar constraint. Hence, it is necessary to consider homograph or add sensors to further exploration.

## VI. CONCLUSION

In this paper, we propose an approximate piecewise constant curvature kinematics equivalent model, i.e. APCC 2L-5R; based on this model, we give an important application, configuration estimation of planar continuum robot through "hand-eye cameras." Through the APCC 2L-5R model, the kinematics of the classical rigid linkages is equivalent to that the real continuum robot in approximate piecewise constant curvature, and the deviation correction is considered to make the model closer to the real continuum robot. In the application case, we installed monocular cameras at the end of each ACC-segment of the continuum robot, and established a generalized polar constraint relationship through the equivalent model, then solved each segment of the continuum robot one by one, and finally obtained the overall configuration estimation. Experiments show that the APCC 2L-5R model is effective and can be effectively applied to environment perception of continuum robots. In the next step, we will use this model to estimate configuration of planar continuum robot in real-time, and further realize effective feedback control. Furthermore, the kinematics equivalent model APCC 2L-5R can be applied to the planning and control of continuum robots, taking equivalent advantage of classical rigid linkages.